\documentclass[twoside,11pt]{article}

\usepackage{blindtext}

\usepackage[preprint]{jmlr2e}

\usepackage{lastpage}
\jmlrheading{23}{2022}{1-\pageref{LastPage}}{1/21; Revised 5/22}{9/22}{21-0000}{Masanari Kimura and Howard Bondell}

\usepackage{amsmath,amsfonts,bm}

\def\eqref#1{equation~\ref{#1}}

\def\1{\bm{1}}

\def\eps{{\epsilon}}

\DeclareMathAlphabet{\mathsfit}{\encodingdefault}{\sfdefault}{m}{sl}
\SetMathAlphabet{\mathsfit}{bold}{\encodingdefault}{\sfdefault}{bx}{n}

\DeclareMathOperator*{\argmax}{arg\,max}
\DeclareMathOperator*{\argmin}{arg\,min}

\usepackage{hyperref}
\usepackage{url}
\usepackage{amsmath}
\usepackage{amssymb}
\usepackage{mathtools}
\usepackage{booktabs}
\usepackage{bbm}
\usepackage{color}

\allowdisplaybreaks

\ShortHeadings{Test-Time Augmentation Meets Variational Bayes}{Kimura and Bondell}
\firstpageno{1}

\begin{document}

\title{Test-Time Augmentation Meets Variational Bayes}

\author{\name Masanari Kimura \email m.kimura@unimelb.edu.au \\
      \addr School of Mathematics and Statistics \\
      The University of Melbourne
      \AND
      \name Howard Bondell \email howard.bondell@unimelb.edu.au \\
      \addr School of Mathematics and Statistics \\
      The University of Melbourne
      }

\editor{My editor}

\maketitle

\begin{abstract}
Data augmentation is known to contribute significantly to the robustness of machine learning models.
In most instances, data augmentation is utilized during the training phase.
Test-Time Augmentation (TTA) is a technique that instead leverages these data augmentations during the testing phase to achieve robust predictions.
More precisely, TTA averages the predictions of multiple data augmentations of an instance to produce a final prediction.
Although the effectiveness of TTA has been empirically reported, it can be expected that the predictive performance achieved will depend on the set of data augmentation methods used during testing.
In particular, the data augmentation methods applied should make different contributions to performance.
That is, it is anticipated that there may be differing degrees of contribution in the set of data augmentation methods used for TTA, and these could have a negative impact on prediction performance.
In this study, we consider a weighted version of the TTA based on the contribution of each data augmentation.
Some variants of TTA can be regarded as considering the problem of determining the appropriate weighting.
We demonstrate that the determination of the coefficients of this weighted TTA can be formalized in a variational Bayesian framework.
We also show that optimizing the weights to maximize the marginal log-likelihood suppresses candidates of unwanted data augmentations at the test phase.
\end{abstract}

\begin{keywords}
  test-time augmentation, data augmentation, variational Bayes
\end{keywords}

\section{Introduction}
Machine learning has been used successfully in many fields, including computer vision~\citep{guo2016deep,voulodimos2018deep,mahadevkar2022review}, natural language processing~\citep{powers2012machine,alshemali2020improving,goldberg2022neural}, and signal processing~\citep{hu2002handbook,francca2021overview}.
It is widely accepted that the performance of such machine learning algorithms depends mainly on the training data.
That is, if a large amount of high-quality training data is obtained, it is expected that a good model can be obtained.
Indeed, in the field of natural language processing, the scaling law is known that as the number of model parameters, the size of the dataset, and the computational resources used for training increase, the loss decreases according to a power law~\citep{kaplan2020scaling}.
However, in many real-world problem settings, there is no guarantee that the quality and quantity of training data are sufficient.
For example, it is often impossible to collect enough training data for tasks with high data observation costs.
As another example, crowdsourcing for annotation to create teacher labels may result in noisy training data due to variations in the quality of the workers.
These examples suggest that the quantity and quality of the training dataset may be insufficient in real-world problem settings.

Therefore, in practical applications, it is necessary to apply some regularization techniques in the training procedure.
One of the most commonly used approaches is data augmentation.
Data augmentation is a framework in which some transformation of a given training instance is included in the new training data.
The usefulness of data augmentation has been reported for various tasks such as classification~\citep{wong2016understanding,perez2017effectiveness,mikolajczyk2018data}, segmentation~\citep{zhao2019data,sandfort2019data,ghiasi2021simple}, image generation~\citep{dong2017i2t2i,tran2021data} and anomaly detection~\citep{lim2018doping,kimura2019anomaly,castellini2023adversarial}.
Examples of simple data augmentation are injecting Gaussian noise into training instances or affine transformations.
More recent studies have also shown that the synthesis of multiple training instances is effective~\citep{zhang2017mixup,yun2019cutmix,chen2022transmix,liu2022tokenmix}.
These techniques are useful as a practical solution when inadequate training data are given.

These data augmentation techniques are typically applied during the training phase.
Indeed, many neural network training strategies based on stochastic gradient descent add augmented instances into the training mini-batches.
However, it has recently been reported that using these techniques in the testing phase can contribute to improving the predictive performance of the model.
This framework is called Test-Time-Augmentation (TTA), and its usefulness in various tasks has been shown~\citep{wang2019automatic,moshkov2020test,kimura2021understanding,kandel2021improving,cohen2024improving}.
The central idea of TTA is to make predictions with a trained model on a set of input instances at test time to which multiple data augmentations have been applied and to make the aggregation of these predictions as the final prediction.
This aggregation is often performed on the outputs of the model in the case of regression problems, or on the softmax outputs of the model in the case of classification problems.
In addition, various variants of TTA have been proposed due to the simplicity of the idea and its extensibility~\citep{lyzhov2020greedy,mocerino2021adaptta,tomar2022opttta,tomar2023tesla,xiong2023stta}.
There are several studies investigating when TTA is effective~\citep{kimura2021understanding,zhang2022memo,conde2023approaching}.
In particular, this study focuses on the effectiveness of TTA in cases where the training data is noisy.
In such a setup, a given training instance has multiple labels that are not consistent.
Considering again the example of crowdsourcing annotations to data, such noisy cases occur frequently due to differences in the skill level of the annotators and the ambiguity of the target instance.
Such noisy training data is often due to annotation quality or instance ambiguity.
If the training instances and labels are not determined one-to-one, it is anticipated that the predictions of the model are negatively affected.
We can conjecture that TTA can be regarded as correcting these predictions at test time by performing aggregation of multiple outputs.

\begin{figure}
    \centering
    \includegraphics[width=0.95\linewidth]{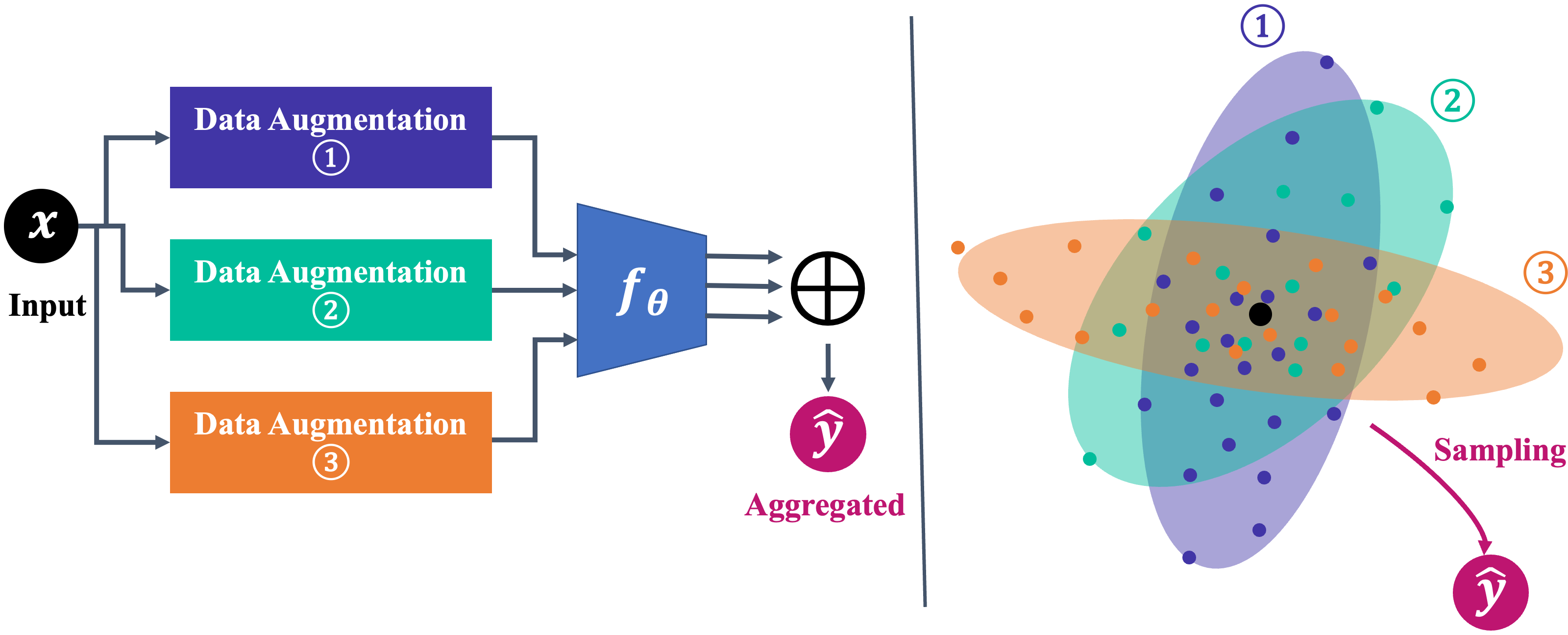}
    \caption{Test-Time Augmentation as Bayesian mixture model. Assuming that the transformed instances acquired by each data augmentation follow some probability distribution, the TTA procedure can be regarded as sampling from their mixture models.}
    \label{fig:vb_tta}
\end{figure}

In this study, we consider formalizing the TTA procedure in noisy training environments using the variational Bayesian framework~\citep{fox2012tutorial,corduneanu2001variational,ghahramani2000propagation}.
We assume that the instances transformed by each data augmentation can be viewed as a perturbation of the test instance drawn from some probability distribution.
The TTA procedure can then be regarded as sampling from a mixture model of these distributions (see Figure~\ref{fig:vb_tta}).
By linking TTA and variational inference in this way, we show that it is possible to weight the set of data augmentation methods used in TTA according to their contributions.
That is, the optimization of weighting coefficients can upweight important data augmentations while suppressing candidates of unnecessary data augmentation during the test phase.

Our contributions are summarized as follows.
\begin{itemize}
    \item We introduce that the TTA framework can be formalized as a Bayesian mixture model. As this framework requires modifications to the formalization depending on whether the predicted labels of the assumed tasks are continuous or discrete, we shall present these in separate subsections.
    \item We show that the optimization of the weighting coefficients through this formalization can suppress unnecessary candidates of data augmentation in the test phase. This suggests that the choice of an appropriate data augmentation strategy, known to be one of the major challenges faced by a typical TTA framework, can be addressed.
    \item Numerical experiments show that our framework allows for appropriate weighting of the TTA procedure. In particular, we demonstrate that in the illustrative examples, the weight coefficients are appropriately evolved by our optimization. In addition, statistical hypothesis tests confirm the validity of the assumptions made by our method, and performance evaluation experiments on real-world data sets demonstrate its effectiveness.
\end{itemize}
Finally, we provide the organization of this paper.
Section~\ref{sec:background} summarizes the background and prior knowledge required for this study.
Section~\ref{sec:vb_tta} introduces the formalization of the TTA procedure as a Bayesian mixture model.
Section~\ref{sec:numerical_experiments} provides the results of numerical experiments to investigate the behavior of our formalization.
To conclude, Section~\ref{sec:conclusion} contains a discussion and possible future works.

\begin{figure}[t]
    \centering
    \includegraphics[width=0.9\linewidth]{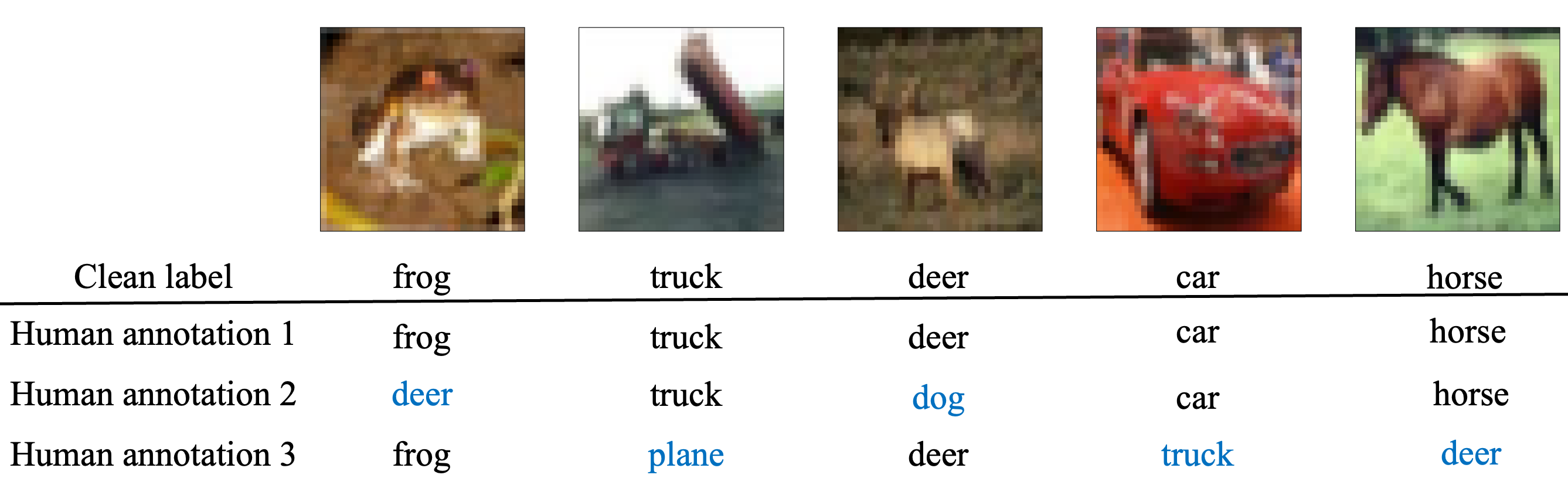}
    \caption{Some example instances in the CIFAR10-N~\citep{wei2021learning} dataset. Each instance in this dataset has three human annotations, which are often inconsistent.}
    \label{fig:cifar10_n}
\end{figure}

\section{Background and Preliminary}
\label{sec:background}
In this work, we consider the supervised learning problem.
Let $\mathcal{X} \subset \mathbb{R}^d$ be a $d$-dimensional input space, and let $\mathcal{Y}$ be an output space.
The goal of supervised learning is to obtain a good model $f_{\bm{\theta}}:\mathcal{X}\to\mathcal{Y}$ parameterized by $\bm{\theta} \in \Theta$, where $\Theta \subset \mathbb{R}^r$ is an $r$-dimensional parameter space.
In the training phase, a training sample $\mathcal{D} = \{(\bm{x}_i, y_i)\}^{N}_{i=1}$ is used to optimize the parameter as
\begin{align}
    \hat{\bm{\theta}} = \argmin_{\bm{\theta} \in \Theta} \frac{1}{N}\sum^N_{i=1}\ell(f_{\bm{\theta}}(\bm{x}_i), y_i), \label{eq:erm}
\end{align}
where $\ell:\mathcal{Y}\times\mathcal{Y}\to[0,\infty)$ is a loss function.
Since $\hat{\bm{\theta}}$ is a minimizer of the empirical risk, it is a consistent estimator for the expected risk under the i.i.d. assumption and is expected to induce correct predictions for unknown data.
This fundamental principle is called Empirical Risk Minimization (ERM)~\citep{vapnik1999overview,vapnik2013nature,hastie2009elements,james2013introduction}.
However, in real-world situations, it is not necessarily true that the model $f_{\hat{\bm{\theta}}}$ obtained in this way leads to ideal predictions.
For example, it is known that under a distribution shift, where the probability distributions followed by the training and test data differ, the estimators obtained by ERM do not satisfy consistency~\citep{shimodaira2000improving,sugiyama2007covariate,quinonero2008dataset,moreno2012unifying}.
Another common case is inconsistent labeling of training data due to variations in annotation quality, sensor failure, and other factors~\citep{patrini2017making,frenay2013classification}.

In this study, we focus specifically on inconsistent labeling settings.
Figure~\ref{fig:cifar10_n} shows examples of several instances in CIFAR10-N~\citep{wei2021learning}, one of the datasets for such problem setting.
To address these issues, we need to consider some modifications or regularizations of the estimator.
One powerful tool for such regularization is data augmentation.
In the following, we summarize the data augmentation used in the training phase and the technique called Test-Time Augmentation, which utilizes them in the testing phase.

\subsection{Data Augmentation}
Data augmentation is a technique in which instances in the training sample $\mathcal{D}$ are transformed in some way to generate new training instances~\citep{van2001art,shorten2019survey,perez2017effectiveness}.
Numerous studies have reported the usefulness of data augmentation, and various data augmentation methods have been developed.
The simplest one is the strategy of generating a new instance $\tilde{\bm{x}} = \bm{x} +\bm{\eps}$ by adding Gaussian noise $\bm{\eps}$ to the training instance $\bm{x}$.
Another common method is to apply a random affine transformation to instance $\bm{x}$ using some $A\in\mathbb{R}^{d\times d}$ and $\bm{b}\in\mathbb{R}^d$ as $\tilde{\bm{x}} = A\bm{x} + \bm{b}$.
More recent studies have shown the effectiveness of a method called mixup~\citep{zhang2017mixup}, in which the convex combination $\tilde{\bm{x}}_{ij} = (1-\lambda)\bm{x}_i + \lambda\bm{x}_j$ of two instances $\bm{x}_i,\bm{x}_{j}\in\mathcal{D}$ with $\lambda \in [0, 1]$ is used as the new training instance~\citep{liang2018understanding,kimura2021mixup,carratino2022mixup}.
Numerous variants have also been proposed due to the simplicity of the idea~\citep{verma2019manifold,chou2020remix,guo2020nonlinear,xu2020adversarial,kim2020puzzle}.
Furthermore, the framework of differentiable automatic data augmentation that performs end-to-end selection and tuning of these data augmentation methods is also attracting attention~\citep{li2020dada,li2020differentiable,hataya2020faster}.

There are also many studies on when data augmentation is effective.
Several studies have reported that data augmentation contributes to improved robustness against out-of-distribution data and adversarial attacks~\citep{rebuffi2021fixing,hendrycks2021many,yao2022improving,volpi2018generalizing}.
The behavior of data augmentation in a setting with label noise is also studied~\citep{nishi2021augmentation,jiang2020beyond}.

\subsection{Test-Time Augmentation}
In general, it is assumed that data augmentation techniques are applied during the training phase.
In recent years, however, there has been a growing interest in utilizing these data augmentation techniques in the testing phase.
This framework is called Test-Time Augmentation (TTA)~\citep{wang2019aleatoric,wang2019automatic,kimura2021understanding}.
Let $\Gamma = \{\varphi_k\}^K_{k=1}$ be a set of data augmentation functions $\varphi_k\colon\mathcal{X}\to\mathcal{X}$.
TTA considers the following prediction using $\Gamma$.
\begin{align}
    \hat{y} = \frac{1}{K}\sum^K_{k=1}(f_{\hat{\bm{\theta}}}\circ\varphi_k)(\bm{x}) = \frac{1}{K}\sum^K_{k=1}f_{\hat{\bm{\theta}}}(\varphi_k(\bm{x})), \label{eq:tta}
\end{align}
where $\bm{x} \in \mathcal{X}$ is an arbitrary input vector given in the test phase, and $\hat{\bm{\theta}}$ is the parameter obtained by Eq.~\ref{eq:erm}.
In recent years, several variants of TTA have been proposed.
\citet{kim2020learning} propose to learn a meta-model that predicts expected loss by applying $\varphi_k \in \Gamma$ and selecting the best candidate.
They report that better performance is achieved by applying the top-$K$ transformations chosen by their method than by randomly applying top-$K$ data augmentations.
\citet{shanmugam2021better} also propose to learn a weighting matrix of aggregated predictions that minimizes validation loss.
Similar ideas have been proposed by \citet{son2023efficient} and \citet{xiong2023stta}, named Selective Test-Time Augmentation.

These variants can be regarded as considering the choice of the coefficients $w_k(\bm{x})$ of the following weighted TTA.
\begin{align}
    \hat{y} = \frac{1}{K}\sum^K_{k=1}w_k(\bm{x})\cdot(f_{\hat{\bm{\theta}}}\circ\varphi_k)(\bm{x}) = \frac{1}{K}\sum^K_{k=1}w_k(\bm{x})\cdot f_{\hat{\bm{\theta}}}(\varphi_k(\bm{x})),
\end{align}
where $\sum^K_{k=1} w_k(\bm{x}) = 1$.
Indeed, the \citet{kim2020learning} method of selecting the top-$K$ transformations such that the expected loss is minimized is equivalent to setting $w_k(\bm{x}) = 0$ on unnecessary candidates.
That is, selecting useful candidates for the testing phase from a set of data augmentations is an essential problem.

\paragraph{Test-Time Augmentation under Noisy Environments}
If the training sample of sufficient size is perfectly clean and from the identical distribution as the test data, then the prediction by the minimizer of the ERM obtained in Eq.~\ref{eq:erm} should be optimal.
However, because of the noisy nature of the real-world problem setting, TTA is expected to be effective.
In particular, we conjecture that TTA is effective in cases where the predictions for the input vector are not uniquely determined in the training data.
In this study, we consider the noisy training sample where each $\bm{x} \in \mathcal{D}$ has a set of noisy labels $S_{\bm{x}}$.
We assume that $S_{\bm{x}}$ includes several inconsistent class labels in the classification problem and $S_{\bm{x}}$ includes several perturbed responses in the regression problem.

\paragraph{Why is Determination of TTA Weight Coefficients Difficult?}
\label{sec:difficulty_determination_tta_weight}
To achieve good prediction, the coefficients $w_k$ of the weighted TTA need to be determined appropriately.
In particular, finding out which data augmentation strategies $\varphi_k$ are unnecessary in TTA is an important issue.
Intuitively, the simplest idea seems to be to eliminate $\varphi_k$ such that the predictive performance of the composite function $f_{\hat{\bm{\theta}}} \circ \varphi_k$ is poor.
However, simple numerical experiments can provide a counterexample to this idea (see Table~\ref{tab:tta_combination}).
The dataset used in this experiment is the MNIST database (Modified National Institute of Standards and Technology database)~\citep{deng2012mnist} and the base model is a simple three-layer convolutional neural network for $\lbrack A \rbrack$ ERM (Empirical Risk Minimization).
Here, the MNIST database is a large collection of handwritten digits, and it has a training set of $60,000$ examples and a test set of $10,000$ examples.
In this experiment, we only use $50$ instances for training of the base model.
Here, the training instances are randomly selected for each trial.
The experimental results are the means and standard deviations of ten trials with different random seeds.
We consider a combination of four data augmentation methods ($\lbrack B \rbrack$ Rotate $20^\circ$, $\lbrack C \rbrack$ Rotate $-20^\circ$, $\lbrack D \rbrack$ Gaussian noise, and  $\lbrack E \rbrack$ Mixup~\citep{zhang2017mixup}) to evaluate the performance of TTA.
In this table, $K$-TTA denotes a TTA based on $K$ data augmentation strategies.
The experiments show that the performance of TTA based on a single data augmentation is relatively good with Gaussian Noise and Mixup, while the two Rotate strategies are poor.
However, considering these combinations, it can be seen that the best performance is achieved when Rotate is included ($4$-TTA ($A + B + C + D$)), which exceeds the performance of those that exclude Rotate ($3$-TTA ($A + D + E$)).
This simple experiment shows that the single worst-performing data augmentation method is not always unnecessary in TTA.
This result motivates the development of a method for determining the coefficients of the weighted TTA by an appropriate procedure.

\begin{table}[t]
    \centering
    \caption{Performance evaluation of TTA with all combinations of data augmentation strategies. Each TTA with a single data augmentation strategy is labeled as $A$, $B$, $C$, $D$, $E$ respectively.
    $K$-TTA is a TTA based on $K$ data augmentation strategies. The experimental results reported in this table are the means and standard deviations of ten trials with different random seeds.}
    \begin{tabular}{l|c}
        \toprule
         TTA Strategy      & Evaluation (Accuracy)   \\
         \midrule
         $\lbrack A \rbrack$ ERM (w/o TTA) &  $0.6210 (\pm 0.052)$   \\
         \hline
         $\lbrack B \rbrack$ $1$-TTA (Rotate $20^\circ$) & $0.4729 (\pm 0.049)$ \\
         $\lbrack C \rbrack$ $1$-TTA (Rotate $-20^\circ$) & $0.4747 (\pm 0.052)$ \\
         $\lbrack D \rbrack$ $1$-TTA (Gaussian noise) & $0.6205 (\pm 0.046)$ \\
         $\lbrack E \rbrack$ $1$-TTA (Mixup~\citep{zhang2017mixup}) & $0.6118 (\pm 0.047)$ \\
         \hline
         $2$-TTA ($A + B$) & $0.6018 (\pm 0.044)$ \\
         $2$-TTA ($A + C$) & $0.6041 (\pm 0.042)$ \\
         $2$-TTA ($A + D$) & $0.6205 (\pm 0.040)$ \\
         $2$-TTA ($A + E$) & $0.6192 (\pm 0.045)$ \\
         \hline
         $3$-TTA ($A + B + C$) & $0.6252 (\pm 0.036)$ \\
         $3$-TTA ($A + B + D$) & $0.6175 (\pm 0.037)$ \\
         $3$-TTA ($A + B + E$) & $0.6147 (\pm 0.031)$ \\
         $3$-TTA ($A + C + D$) & $0.6252 (\pm 0.033)$ \\
         $3$-TTA ($A + C + E$) & $0.6205 (\pm 0.029)$ \\
         $3$-TTA ($A + D + E$) & $0.6201 (\pm 0.033)$ \\
         \hline
         $4$-TTA ($A + B + C + D$) & ${\bf 0.6356 (\pm 0.022)}$ \\
         $4$-TTA ($A + B + C + E$) & $0.6328 (\pm 0.024)$ \\
         $4$-TTA ($A + B + D + E$) & $0.6211 (\pm 0.024)$ \\
         $4$-TTA ($A + C + D + E$) & $0.6276 (\pm 0.022)$ \\
         \hline
         $5$-TTA ($A + B + C + D + E$) & ${\bf 0.6355 (\pm 0.020)}$ \\
         \bottomrule
    \end{tabular}
    \label{tab:tta_combination}
\end{table}

\section{Test-Time Augmentation as Bayesian Mixture Model}
\label{sec:vb_tta}
In this section, we consider the reformulation of the TTA procedure as a Bayesian mixture model.
The key idea is to regard the final output obtained by the TTA procedure as being acquired by sampling from a mixture of distributions made by several data augmentation functions.
The following discussion is divided according to whether the prediction problem is a continuous or categorical case.
Here, we first consider the continuous case (Section.~\ref{subsec:vb_tta_continuous}) as the formulation is most straightforward and can then be extended to discuss the categorical case (Section.~\ref{subsec:vb_tta_categorical}).

\subsection{Continuous Case}
\label{subsec:vb_tta_continuous}
We consider $Y = f_{\bm{\theta}}(X) + \epsilon$, where $\epsilon \sim \mathcal{N}(0, \sigma_{\epsilon})$.
We also assume that the transformation $\bm{\xi}_{k,\bm{x}} \coloneqq \varphi_k(\bm{x})$ of an instance $\bm{x}$ by a data augmentation $\varphi_k \in \Gamma$ follows a Gaussian distribution as $\bm{\xi}_{k,\bm{x}}\sim\mathcal{N}(\bm{x}, \Sigma_k)$, where $\Sigma_k \in \mathbb{R}^{d\times d}$.
That is, the mapping by a given data augmentation is assumed to be normally distributed around the original instance.
The Taylor expansion of $f_{\bm{\theta}}(\bm{\xi}_{k,\bm{x}})$ around $\bm{x}$ yields
\begin{align}
    \mu_k(\bm{x};\bm{\theta}) &\coloneqq \mathbb{E}_{\bm{\xi}_{k,\bm{x}}}\left[f_{\bm{\theta}}(\bm{\xi}_{k,\bm{x}})\right] \approx f_{\bm{\theta}}(\bm{x}), \\
    \Sigma_k(\bm{x};\bm{\theta}) &\coloneqq \mathbb{V}\left[f_{\bm{\theta}}(\bm{\xi}_{k,\bm{x}})\right] \nonumber \\
    &\approx \frac{1}{N}\nabla f^\top_{\bm{\theta}}(\bm{\xi}_{k,\bm{x}})\mathrm{Cov}\left(\bm{\xi}_{k,\bm{x}}\right)\nabla f_{\bm{\theta}}(\bm{\xi}_{k,\bm{x}}) + \sigma_\epsilon \nonumber \\
    &= \frac{1}{N} \nabla f^\top_{\bm{\theta}}(\varphi_k(\bm{x})) \Sigma_k \nabla f_{\bm{\theta}}(\varphi_k(\bm{x})) + \sigma_\epsilon.
\end{align}
That is, we can say that
\begin{align}
    \sqrt{N}\left(f_{\bm{\theta}}\left(\frac{1}{N}\sum^N_{j=1}\bm{\xi}_{{k,\bm{x}}_j}\right) - f_{\bm{\theta}}(\bm{x})\right) \rightsquigarrow \mathcal{N}\left(0, \left(\frac{\partial}{\partial \bm{x}}f_{\bm{\theta}}(\bm{x})\right)^\top\Sigma_k \left(\frac{\partial}{\partial \bm{x}}f_{\bm{\theta}}(\bm{x})\right) + \sigma_{\epsilon_i} \right),
\end{align}
where the symbol $\rightsquigarrow$ stands for the convergence in distribution (or convergence in law).
Then, we have $f_{\bm{\theta}}(\varphi_k(\bm{x})) \sim \mathcal{N}(\mu_k(\bm{x};\bm{\theta}), \Sigma_k(\bm{x};\bm{\theta}))$ and
\begin{align}
    P(y \mid \bm{x}, \bm{w}, \Sigma_k) &= \sum^K_{k=1}w_k\cdot\mathcal{N}(y \mid \mu_k(\bm{x};\bm{\theta}), \Sigma_k(\bm{x};\bm{\theta})), \\
    P(S_{\bm{x}} \mid \bm{x}, \bm{w}, \Sigma_k) &= \prod^{|S_{\bm{x}}|}_{j=1}\sum^K_{k=1}w_k\cdot\mathcal{N}(y_j \mid \mu_k(\bm{x};\bm{\theta}), \Sigma_k(\bm{x};\bm{\theta})), \label{eq:distribution_observed_data}
\end{align}
where $\bm{w} = \{w_k\}^K_{k=1}$.
Now consider a binary random variable $\bm{z} = \{z_{jk}\}$ and set $z_{jk}=1$ only when a label $y_j$ is generated from the $k$-th distribution.
Using this variable, we can write as
\begin{align}
    P(S_{\bm{x}} \mid \bm{x}, \bm{z}, \Sigma_k) = \prod^{|S_{\bm{x}}|}_{j=1}\prod^K_{k=1}\mathcal{N}(y_j \mid \mu_k(\bm{x};\bm{\theta}),\Sigma_k(\bm{x};\bm{\theta}))^{z_{jk}}, \label{eq:distribution_with_binary}
\end{align}
and,
\begin{align}
    P(\bm{z} \mid \bm{w}) = \prod^K_{k=1}\prod^{S_{\bm{x}}}_{j=1}z_{jk}. \label{eq:distribution_pi}
\end{align}
By marginalizing Eq.~\ref{eq:distribution_with_binary} with the prior distribution of Eq.~\ref{eq:distribution_pi}, the marginal distribution of the observed data of Eq.~\ref{eq:distribution_observed_data} can be recovered.

We consider the following conjugate priors over $\mu_{\bm{x},\bm{\theta}} = \{\mu_k(\bm{x};\bm{\theta})\}$ and $\Sigma_{\bm{x},\bm{\theta}} = \{\Sigma_k(\bm{x};\bm{\theta})\}$.
\begin{align}
    P(\mu_{\bm{x},\bm{\theta}}) &= \prod^K_{k=1}\mathcal{N}(\mu_k(\bm{x};\bm{\theta}) \mid 0, \beta I), \\
    P(\Sigma_{\bm{x},\bm{\theta}}) &= \prod^K_{k=1}\mathcal{W}(\Sigma_k(\bm{x};\bm{\theta}) \mid \nu, V),
\end{align}
where $\beta$ is a fixed and large value that corresponds to a broad prior distribution over $\mu_{\bm{x},\bm{\theta}}$, $I$ is the identity matrix, and $\mathcal{W}$ is the Wishart distribution with $\nu$ degrees of freedom and a scale matrix $V$ having the following density function
\begin{align}
    p(\Sigma) = \frac{|\Sigma|^{(\nu-c-1)/2} \exp\{-\frac{1}{2}\mathrm{tr}(V^{-1}\Sigma)\}}{2^{c\nu/2}z^{c(c-1)/4}|V|^{\nu/2}\prod^c_{i=1}\Gamma\left(\frac{\nu - i + 1}{2}\right)}.
\end{align}
Here, $\Gamma(z) = \int^\infty_0 t^{z-1}\exp(-t)dt$ is the Gamma function.
The joint distribution of all of the random variables conditioned on the weighting coefficients is given by
\begin{align}
    P(S_{\bm{x}}, \mu_{\bm{x},\bm{\theta}}, \Sigma_{\bm{x},\bm{\theta}}, \bm{z} \mid \bm{w}) = P(S_{\bm{x}} \mid \mu_{\bm{x},\bm{\theta}}, \Sigma_{\bm{x}, \bm{\theta}}, \bm{z})P(\bm{z} \mid \bm{w})P(\mu_{\bm{x}, \bm{\theta}})P(\Sigma_{\bm{x},\bm{\theta}}). \label{eq:joint_distribution}
\end{align}

Evaluating $P(S_{\bm{x}} \mid \bm{w})$ requires marginalization of Eq.~\ref{eq:joint_distribution} with respect to $\bm{z}$, $\mu_{\bm{x},\bm{\theta}}$, and $\Sigma_{\bm{x},\bm{\theta}}$, which is analytically intractable.
Therefore, we utilize the variational method to obtain a tractable lower bound for $P(S_{\bm{x}} \mid \bm{w})$.
Denote $\bm{\eta} = \{\bm{z}$, $\mu_{\bm{x},\bm{\theta}}$, $\Sigma_{\bm{x},\bm{\theta}}\}$.
Then the marginal likelihood we need to evaluate is given by
\begin{align}
    P(S_{\bm{x}} \mid \bm{w}) = \int P(S_{\bm{x}}, \bm{\eta} \mid \bm{w})d\bm{\eta}.
\end{align}
In the variational inference framework, we introduce a distribution $Q(\bm{\eta})$ that provides an approximation of the true posterior distribution.
Consider the following transformation of the log marginal likelihood using the distribution $Q(\bm{\eta})$.
\begin{align}
    \ln P(S_{\bm{x}} \mid \bm{w}) &= \ln \int P(S_{\bm{x}}, \bm{\eta} \mid \bm{w})d\bm{\eta} \nonumber \\
    &= \ln \int Q(\bm{\eta})\frac{P(S_{\bm{x}}, \bm{\eta} \mid \bm{w})}{Q(\bm{\eta})}d\bm{\eta} \nonumber \\
    &\geq \int Q(\bm{\eta})\ln\frac{P(S_{\bm{x}}, \bm{\eta} \mid \bm{w})}{Q(\bm{\eta})}d\bm{\eta}.\quad\quad (\because \text{Jensen’s inequality}). \label{eq:elbo}
\end{align}
Here we denote the lower bound by $\mathcal{L}(Q)$.
It is known that a judicious choice of distribution $Q$ makes the quantity $\mathcal{L}(Q)$ tractable to compute, even if the original log-likelihood function is not.
Here, the difference between the true log marginal likelihood $\ln P(S_{\bm{x}} \mid \bm{w})$ and the bound $\mathcal{L}(Q)$ is given by the Kullback--Leibler (KL)-divergence
\begin{align}
    D_{\mathrm{KL}}[Q \| P] = -\int Q(\bm{\eta}) \ln \frac{P(\bm{\eta} \mid S_{\bm{x}}, \bm{w})}{Q(\bm{\eta})}d\bm{\eta}.
\end{align}
The goal of the variational inference framework is to choose an appropriate form of $Q$ that is simple enough to easily evaluate the lower bound $\mathcal{L}(Q)$, but flexible enough to make the lower bound reasonably tight.
Since the true log-likelihood is independent of $Q$, we see that this procedure is equivalent to minimizing the KL-divergence.
Although it is known that $D_{\mathrm{KL}}[Q\|P]=0$ when $P=Q$, for efficient approximation we consider restricting the class of $Q$.

We consider a family of constrained variational distributions, assuming that $Q(\bm{\eta})$ is factorized over a subset $\{\bm{\eta}_i\}$ of the variables in $\bm{\eta}$ as
\begin{align}
    Q(\bm{\eta}) = \prod_{i=1}Q_i(\bm{\eta}_i).
\end{align}
Here, the KL-divergence can be minimized over all possible factorial distributions, and the solutions are given as
\begin{align}
    Q_i(\bm{\eta}_i) = \frac{\exp\{\langle \ln P(S_{\bm{x}}, \bm{\eta})\rangle_{k\neq i}\}}{\int\exp\{\langle \ln P(S_{\bm{x}}, \bm{\eta})\rangle_{k\neq i}\}d\bm{\eta}_i},
\end{align}
where $\langle \cdot \rangle_{k\neq i}$ is an expectation with respect to $Q_i(\bm{\eta}_i)$ for all $k \neq i$.
Then, the variational posterior distributions we are interested in are
\begin{align}
    Q(\bm{\eta}) = Q(\bm{z}, \mu_{\bm{x},\bm{\theta}}, \Sigma_{\bm{x},\bm{\theta}}) = Q_{\bm{z}}(\bm{z})Q_{\mu_{\bm{x},\bm{\theta}}}(\mu_{\bm{x},\bm{\theta}})Q_{\Sigma_{\bm{x},\bm{\theta}}}(\Sigma_{\bm{x}, \bm{\theta}}),
\end{align}
and the solutions for the factors of the variational posterior are
\begin{align}
    Q_{\bm{z}}(\bm{z}) &= \prod^N_{i=1}\prod^{|S_{\bm{x}_i}|}_{j=1}\prod^K_{k=1}p_{jk}^{z_{jk}}, \\
    Q_{\mu_{\bm{x},\bm{\theta}}}(\mu_{\bm{x}, \bm{\theta}}) &= \prod^K_{k=1}\mathcal{N}(\mu_k(\bm{x};\bm{\theta}) \mid m_{k}(\mu_{\bm{x},\bm{\theta}}), G_k(\mu_{\bm{x}, \bm{\theta}})), \\
    Q_{\Sigma_{\bm{x}, \bm{\theta}}}(\Sigma_{\bm{x}, \bm{\theta}}) &= \prod^K_{k=1}\mathcal{W}(\Sigma_k(\bm{x}; \bm{\theta}) \mid \nu_k(\Sigma_{\bm{x},\bm{\theta}}), V_k(\Sigma_{\bm{x},\bm{\theta}})),
\end{align}
where
\begin{align}
    p_{jk} &= \frac{\tilde{p}_{jk}}{\sum^K_{k=1}\tilde{p}_{jk}}, \\
    \tilde{p}_{jk} &= \exp\left\{\frac{1}{2}\mathbb{E}[\ln \Sigma_k(\bm{x}; \bm{\theta})] + \ln w_k -\frac{1}{2}\mathrm{tr}\left(\mathbb{E}[\Sigma_k(\bm{x},\bm{\theta})](y_j -\mathbb{E}[\mu_k(\bm{x};\bm{\theta})])^2\right) \right\}, \\
    m_k(\mu_{\bm{x}, \bm{\theta}}) &= G_k^{-1}(\mu_{\bm{x}, \bm{\theta}})\mathbb{E}[\Sigma_k( \bm{x}; \bm{\theta})]\sum^{|S_{\bm{x}}|}_{j=1}y_j\mathbb{E}[z_{jk}], \\
    G_k(\mu_{\bm{x},\bm{\theta}}) &= \beta I + \mathbb{E}[\Sigma_k(\bm{x}; \bm{\theta})]\sum^{|S_{\bm{x}}|}_{j=1}\mathbb{E}[z_{jk}], \\
    \nu_k(\Sigma_{\bm{x}, \bm{\theta}}) &= \nu + \sum^{|S_{\bm{x}}|}_{j=1}\mathbb{E}[z_{jk}], \\
    V_k(\Sigma_{\bm{x}, \bm{\theta}}) &= V + \sum^{|S_{\bm{x}}|}_{j=1}y_j^2\mathbb{E}[z_{jk}] - \mathbb{E}[\mu_k(\bm{x}; \bm{\theta})]\sum^{|S_{\bm{x}}|}_{j=1}y_j\mathbb{E}[z_{jk}] \nonumber \\
    &\quad\quad\quad - \left(\sum^{|S_{\bm{x}}|}_{j=1}y_j\mathbb{E}[z_{jk}]\right)\mathbb{E}[\mu_k(\bm{x}; \bm{\theta})]^\top + \mathbb{E}[\mu_k(\bm{x};\bm{\theta})^2]\sum^{|S_{\bm{x}}|}_{j=1}\mathbb{E}[z_{jk}].
\end{align}

We can evaluate the variational lower bound in Eq.~\ref{eq:elbo} as
\begin{align}
    \mathcal{L}(Q) &= \int Q(\bm{\eta})\ln\frac{P(S_{\bm{x}}, \bm{\eta} \mid \bm{w})}{Q(\bm{\eta})}d\bm{\eta} \nonumber \\
    &= \int Q_{\bm{z}}(\bm{z})Q_{\mu_{\bm{x}, \bm{\theta}}}(\mu_{\bm{x}, \bm{\theta}})Q_{\Sigma_{\bm{x}, \bm{\theta}}}(\Sigma_{\bm{x}, \bm{\theta}}) \nonumber \\
    &\quad\quad\quad \times\ln\frac{P(S_{\bm{x}} \mid \bm{z}, \mu_{\bm{x}, \bm{\theta}}, \Sigma_{\bm{x}, \bm{\theta}})P(\bm{z})P(\mu_{\bm{x}, \bm{\theta}})P(\Sigma_{\bm{x}, \bm{\theta}})}{Q_{\bm{z}}(\bm{z})Q_{\mu_{\bm{x}, \bm{\theta}}}(\mu_{\bm{x}, \bm{\theta}})Q_{\Sigma_{\bm{x}, \bm{\theta}}}(\Sigma_{\bm{x}, \bm{\theta}})}d\bm{z}d\mu_{\bm{x}, \bm{\theta}}d\Sigma_{\bm{x}, \bm{\theta}} \nonumber \\
    &= J_{S_{\bm{x}}\mid\bm{\eta}} + J_{\bm{z}} + J_{\mu_{\bm{x}, \bm{\theta}}} + J_{\Sigma_{\bm{x}, \bm{\theta}}} - \Pi_{\bm{z}} - \Pi_{\mu_{\bm{x}, \bm{\theta}}} - \Pi_{\Sigma(\bm{x}, \bm{\theta})},
\end{align}
where
\begin{align}
    J_{S_{\bm{x}}\mid\bm{\eta}} &\coloneqq \mathbb{E}_{Q}[\ln P(S_{\bm{x}} \mid \bm{\eta})] \nonumber \\
    &= \sum^N_{i=1}\sum^K_{k=1}\sum^{|S_{\bm{x}_i}|}_{j=1}p_{jk}\Bigg(\frac{1}{2}\mathbb{E}[\ln \Sigma_k(\bm{x}; \bm{\theta})] - \frac{c}{2}\ln(2z) \nonumber \\
    &\quad\quad\quad\quad\quad\quad\quad\quad\quad - \frac{1}{2}\mathrm{tr}\left(\mathbb{E}[\Sigma_k(\bm{x}; \bm{\theta})](y_j - \mathbb{E}[\mu_k(\bm{x}; \bm{\theta})])^2\right)\Bigg) \\
    J_{\bm{z}} &\coloneqq \mathbb{E}_{Q_{\bm{z}}}[\ln P(\bm{z})] \nonumber \\
    &= \sum^N_{i=1}\sum^K_{k=1}\sum^{|S_{\bm{x}_i}|}_{j=1}p_{jk}\ln w_k, \\
    J_{\mu_{\bm{x}, \bm{\theta}}} &\coloneqq \mathbb{E}_{Q_{\mu_{\bm{x}, \bm{\theta}}}}[\ln P(\mu_{\bm{x}, \bm{\theta}})] \nonumber \\
    &= Kc\cdot \ln\left(\frac{\beta}{2z}\right) - \frac{\beta}{2}\sum^K_{k=1}\left(G_k(\mu_{\bm{x}, \bm{\theta}})^{-1} + m_k(\mu_{\bm{x}, \bm{\theta}})m_k^\top(\mu_{\bm{x}, \bm{\theta}})\right), \\
    J_{\Sigma_{\bm{x}, \bm{\theta}}} &\coloneqq \mathbb{E}_{Q_{\Sigma_{\bm{x}, \bm{\theta}}}}[\ln P(\Sigma_{\bm{x}, \bm{\theta}})] \nonumber \\
    &=  K\left(-\frac{\nu c}{2}\ln 2 - \frac{c(c-1)}{4}\ln z - \sum^c_{i=1}\ln\Gamma\left(\frac{\nu + 1 - i}{2} \right) + \frac{\nu}{2}\ln |V|\right) \nonumber\\
    &\quad\quad\quad\quad\quad\quad + \frac{\nu - c - 1}{2}\sum^K_{k=1}\mathbb{E}[\ln \Sigma_k(\bm{x}; \bm{\theta})] - \frac{1}{2}\mathrm{tr}\left(V\sum^K_{k=1}\nu_k(\Sigma_{\bm{x}, \bm{\theta}})V^{-1}_k(\Sigma_{\bm{x}, \bm{\theta}})\right), \\
    \Pi_{\bm{z}} &\coloneqq \mathbb{E}_{Q_{\bm{z}}}[\ln Q_{\bm{z}}(\bm{z})] \nonumber \\
    &= \sum^N_{i=1}\sum^K_{k=1}\sum^{|S_{\bm{x}_i}|}_{j=1}p_{jk}\ln p_{jk}, \\
    \Pi_{\mu_{\bm{x}, \bm{\theta}}} &\coloneqq \mathbb{E}_{Q_{\mu_{\bm{x}, \bm{\theta}}}}[\ln Q_{\mu_{\bm{x}, \bm{\theta}}}(\mu_{\bm{x}, \bm{\theta}})] \nonumber \\
    &= \sum^K_{k=1}\left(-\frac{c}{2}(1 + \ln(2z)) + \frac{1}{2}\ln\left|G_k(\mu_{\bm{x}, \bm{\theta}})\right| \right), \\
    \Pi_{\Sigma_{\bm{x}, \bm{\theta}}} &\coloneqq \mathbb{E}_{Q_{\Sigma_{\bm{x}, \bm{\theta}}}}[\ln Q_{\Sigma_{\bm{x}, \bm{\theta}}}(\Sigma_{\bm{x}, \bm{\theta}})] \nonumber \\
    &= \sum^K_{k=1}\Bigg\{-\frac{c\nu_k(\Sigma_{\bm{x}, \bm{\theta}})}{2}\ln 2 - \frac{c(c-1)}{4}\ln z \nonumber \\
    &\quad\quad\quad\quad\quad\quad - \sum^c_{i=1}\ln\Gamma\left(\frac{\nu_k(\Sigma_{\bm{x}, \bm{\theta}}) + 1 - i}{2}\right) + \frac{\nu_k(\Sigma_{\bm{x}, \bm{\theta}})}{2}\ln\left|V_k(\Sigma_{\bm{x}, \bm{\theta}})\right| \nonumber\\
    &\quad\quad\quad\quad\quad\quad + \frac{\nu_k(\Sigma_{\bm{x}, \bm{\theta}}) - c - 1}{2}\mathbb{E}[\ln \Sigma_k(\bm{x}; \bm{\theta})] - \frac{1}{2}\mathrm{tr}\left(V_k(\Sigma_{\bm{x}, \bm{\theta}})\right)\nu_k(\Sigma_{\bm{x}, \bm{\theta}})V^{-1}_k(\Sigma_{\bm{x}, \bm{\theta}})\Bigg\}.
\end{align}
Our aim is to maximize this variational lower bound with respect to $\bm{w}$ to obtain an optimal choice of weighting coefficients.
To do so, we utilize an EM procedure with i) M-step: maximization of $\mathcal{L}(Q)$ with respect to $\bm{w}$, and ii) E-step: update $Q_{\bm{z}}, Q_{\mu_{\bm{x}, \bm{\theta}}}, Q_{\Sigma_{\bm{x}, \bm{\theta}}}$.
We call our framework Variational Bayesian Test-Time Augmentation (VB-TTA).

\subsection{Categorical Case}
\label{subsec:vb_tta_categorical}
In the categorical case, the model given in Eq.~\ref{eq:tta} no longer holds.
To proceed in this case, we consider the following latent variable multinomial probit model as
\begin{align}
    Y &= \argmax Z, \\
    Z &= \begin{pmatrix}
        Z_1 \\ \vdots \\ Z_C
    \end{pmatrix} \sim \mathcal{N}(\bm{\mu}_k(\bm{x}; \bm{\theta}), \Sigma_k(\bm{x}; \bm{\theta})), \\
    \mu_{k,i}(\bm{x}; \bm{\theta}) &\coloneqq \mathbb{E}_{\bm{\xi}_{k,\bm{x}}}\left[f_{\bm{\theta},i}(\bm{\xi}_{k,\bm{x}})\right] \approx f_{\bm{\theta},i}(\bm{x}), \quad \forall 1 \leq i \leq c, \\
    \Sigma_{k,i}(\bm{x}; \bm{\theta}) &\coloneqq \mathbb{V}\left[f_{\bm{\theta},i}(\bm{\xi}_{k,\bm{x}})\right] \approx \frac{1}{N}\left(\frac{\partial}{\partial \bm{x}}f_{\bm{\theta},i}(\bm{x})\right)^\top\Sigma_k \left(\frac{\partial}{\partial \bm{x}}f_{\bm{\theta},i}(\bm{x})\right) + \sigma_{\epsilon_i}, \quad \forall 1 \leq i \leq c.
\end{align}
Here, we assume that $f_{\bm{\theta}}$ is a vector-valued function and $f_{\bm{\theta}, i}(\bm{x})$ is the $i$-th element of $f_{\bm{\theta}}(\bm{x})$.
Then, for all $1 \leq i \leq C$,
\begin{align}
    Z_i \sim \mathcal{N}(\mu_{k,i}(\bm{x}; \bm{\theta}), \sigma_{k,i}(\bm{x}; \bm{\theta})) = \mathcal{N}\left(f_{\bm{\theta},i}(\bm{x}), \frac{1}{N}\left(\frac{\partial}{\partial \bm{x}}f_{\bm{\theta},i}(\bm{x})\right)^\top\Sigma_k \left(\frac{\partial}{\partial \bm{x}}f_{\bm{\theta},i}(\bm{x})\right) + \sigma_{\epsilon_i} \right),
\end{align}
where $C$ is the number of classes.
Here, we can see that $Y$ has the probability mass function
\begin{align}
    P(Y = i) &= P(Z_i = \max_{j} Z_j)
\end{align}

For example, consider the probability that $Y=1$ with $C=3$.
In this case,
\begin{align}
    P(Y=1) &= P(Z_1 > Z_2 \cap Z_1 > Z_3) \nonumber \\
    &= \mathbb{E}\left[P(Z_1 > Z_2 \cap Z_1 > Z_3 \mid Z_2, Z_3)\right] \nonumber \\
    &= \mathbb{E}\left[P(Z_1 > \max(Z_2, Z_3) \mid Z_2, Z_3)\right] \nonumber \\
    &= \mathbb{E}\left[1 - \Phi\left(\frac{\max(Z_2, Z_3) - \mu^*_1}{\sigma^*_1}\right)\right] \nonumber \\
    &= \int \left\{1 - \Phi\left(\frac{\max(Z_2, Z_3) - \mu^*_1}{\sigma^*_1}\right)\right\} p(\max(Z_2, Z_3)) d\max(z_2, z_3).
\end{align}
Here, $\Phi(\cdot)$ is the cumulative distribution function of the standard normal distribution as
\begin{align}
    \Phi(z) = \frac{1}{\sqrt{2z}}\int^z_{-\infty}\exp\left\{-\frac{u^2}{2}\right\}du.
\end{align}
For $V = \max(Z_2,\dots,Z_C)$ with $Z_i \sim \mathcal{N}(\mu^*_i, \sigma^*_i)$,
\begin{align}
    F_{i,V}(v) = P(\max(Z_2,\dots,Z_K) \leq v) = \prod^K_{k=1}F_{Z_i}(v),
\end{align}
and
\begin{align}
    p_j(v) = \frac{\partial}{\partial v}\prod^C_{i = 1}F_{Z_i}(v) &= \frac{\partial}{\partial v}\prod^C_{i=1}\Phi\left(\frac{v - \mu^*_i}{\sigma^*_i}\right) \nonumber\\
    &= \frac{\partial}{\partial v}\exp\left\{\sum^C_{i \neq j}\ln \Phi\left(\frac{v - \mu^*_i}{\sigma^*_i}\right)\right\} \nonumber \\
    &= \left(\sum^C_{i \neq j}\frac{1}{\sigma^*_i}\frac{\phi((v - \mu^*_i) / \sigma^*_i)}{\Phi((v - \mu^*_i) / \sigma^*_i)}\right)\exp\left\{\sum^C_{i \neq j}\ln\Phi\left(\frac{v - \mu^*_i}{\sigma^*_i}\right)\right\} \nonumber \\
    &= \left(\sum^C_{i \neq j}\frac{1}{\sigma^*_i}\frac{\phi((v - \mu^*_i) / \sigma^*_i)}{\Phi((v - \mu^*_i) / \sigma^*_i)}\right)\prod^C_{i \neq j}\Phi\left(\frac{v - \mu^*_i}{\sigma^*_i}\right), \nonumber \\
\end{align}
where $\phi$ is the probability density function of the standard normal distribution as
\begin{align}
    \phi(z) = \frac{1}{\sqrt{2z}}\exp\left\{-\frac{1}{2}z^2\right\}.
\end{align}
Then, for $C=3$ case, we have
\begin{align}
    P(Y = 1) = \int_V \left\{1 - \Phi\left(\frac{v - \mu^*_1}{\sigma^*_1}\right)\right\}\left(\sum^3_{i \neq 1}\frac{1}{\sigma^*_i}\frac{\phi((v - \mu^*_i) / \sigma^*_i)}{\Phi((v - \mu^*_i) / \sigma^*_i)}\right)\prod^3_{i \neq 1}\Phi\left(\frac{v - \mu^*_i}{\sigma^*_i}\right) dv.
\end{align}
More generally, we have
\begin{align}
    P(Y = j) = \int_V \left\{1 - \Phi\left(\frac{v - \mu^*_j}{\sigma^*_j}\right)\right\}\left(\sum^C_{i \neq j}\frac{1}{\sigma^*_i}\frac{\phi((v - \mu^*_i) / \sigma^*_i)}{\Phi((v - \mu^*_i) / \sigma^*_i)}\right)\prod^C_{i \neq j}\Phi\left(\frac{v - \mu^*_i}{\sigma^*_i}\right) dv.
\end{align}
Finally, as in the continuous case, we have
\begin{align}
    P(S_{\bm{x}} \mid \bm{x}, \bm{w}, \Sigma_k) &= \prod^{|S_{\bm{x}}|}_{j=1}\sum^K_{k=1}w_k \cdot P(Y = j) \nonumber \\
    &= \prod^{|S_{\bm{x}}|}_{j=1}\sum^K_{k=1}w_k \cdot \int_V \left\{1 - \Phi\left(\frac{v - \mu^*_j}{\sigma^*_j}\right)\right\} \nonumber \\
    &\quad\quad\quad\quad\quad \times \left(\sum^C_{i \neq j}\frac{1}{\sigma^*_i}\frac{\phi((v - \mu^*_i) / \sigma^*_i)}{\Phi((v - \mu^*_i) / \sigma^*_i)}\right)\prod^C_{i \neq j}\Phi\left(\frac{v - \mu^*_i}{\sigma^*_i}\right) dv.
\end{align}
The optimization is then achieved by using this density to evaluate and differentiate the variational lower bound.

\subsection{Automatic Differentiation Variational Inference for VB-TTA}
Although analytical derivation is possible, a unified implementation is desirable for the introduction of more complex prior distributions and for more advanced problem settings (e.g. ordered outputs such as learning to rank problems), since the posterior distributions are not tractable in such cases.
Numerical implementations with the automatic differentiation variational inference (ADVI)~\citep{kucukelbir2017automatic}, are therefore useful.
The key idea of the ADVI is to transform the latent variables of the model into a common space and variational
approximation in the common space.
In this framework, we that the variational distribution is parametrized as $Q(\bm{\eta} \mid \bm{\kappa})$ by a parameter vector $\bm{\kappa}$, and consider the mapping
\begin{align}
    T \colon \mathrm{supp}(P(\bm{\eta})) \to \mathbb{R}^{|\bm{\eta}|},
\end{align}
where $|\bm{\eta}|$ is the dimension of the vector $\bm{\eta}$, and identify the transformed variable $\bm{\zeta} = T(\bm{\eta})$.
The transformed joint distribution $P(S_{\bm{x}}, \bm{\zeta})$ is a function of $\bm{\zeta}$, and has the reparametrization
\begin{align}
    P(S_{\bm{x}}, \bm{\zeta}) = P(S_{\bm{x}}, T^{-1}(\bm{\zeta})) \left| \det J_{T^{-1}}(\bm{\zeta}) \right|,
\end{align}
where $J_{T^{-1}}(\bm{\zeta})$ is the Jacobian of the inverse of the mapping $T$.

After the transformation, we utilize a full-rank Gaussian variational approximation for $Q$ with $\bm{\kappa} = (\bm{\mu}_{\bm{\zeta}}, \bm{\Sigma}_{\bm{\zeta}})$ as
\begin{align}
    Q(\bm{\zeta} \mid \bm{\kappa}) = \mathcal{N}(\bm{\zeta} \mid \bm{\mu}_{\bm{\zeta}}, \bm{\Sigma}_{\bm{\zeta}}).
\end{align}
To ensure that $\bm{\Sigma}_{\bm{\zeta}}$ always remains positive semidefinite, covariance matrix is reparameterize by using a Cholesky decomposition $\bm{\Sigma} = \bm{L}\bm{L}^\top$~\citep{pinheiro1996unconstrained}, where $\bm{L}$ is a real lower triangular matrix with positive diagonal entries.
Then, the full-rank Gaussian becomes $Q(\bm{\zeta} \mid \bm{\kappa}) = \mathcal{N}(\bm{\zeta} \mid \bm{\mu}_{\bm{\zeta}}, \bm{L}\bm{L}^\top)$.
The ADVI is done by computing the following transformed variational lower bound:
\begin{align}
    \int Q(\bm{\eta})\ln\frac{P(S_{\bm{x}},\bm{\eta})}{Q(\bm{\eta})}d\bm{\eta} &= \int Q(\bm{\zeta} \mid \bm{\kappa}) \ln \left\{\frac{P(S_{\bm{x}}, T^{-1}(\bm{\zeta})\left|\det J_{T^{-1}}(\bm{\zeta})\right|)}{Q(\bm{\zeta} ; \bm{\kappa})}\right\}d\bm{\zeta} \nonumber \\
    &= \int Q(\bm{\zeta} \mid \bm{\kappa}) \ln \left\{P(S_{\bm{x}}, T^{-1}(\bm{\zeta}))\left|\det J_{T^{-1}}(\bm{\zeta})\right| \right\}d\bm{\zeta} \nonumber \\
    &\quad\quad\quad\quad - \int Q(\bm{\zeta} \mid \bm{\kappa}) \ln Q(\bm{\zeta} \mid \bm{\kappa})d\bm{\zeta} \nonumber \\
    &= \mathbb{E}_{Q(\bm{\zeta} \mid \bm{\kappa})}\left[\ln P(S_{\bm{x}} \mid T^{-1}(\bm{\zeta})) + \ln \left|\det J_{T^{-1}}(\bm{\zeta})\right|\right] - \mathbb{E}_{Q(\bm{\zeta} \mid \bm{\kappa})}\left[\ln Q(\bm{\zeta} \mid \bm{\kappa})\right] \nonumber \\
    &= \mathbb{E}_{Q(\bm{\zeta} \mid \bm{\kappa})}\left[\ln P(S_{\bm{x}} \mid T^{-1}(\bm{\zeta})) + \ln \left|\det J_{T^{-1}}(\bm{\zeta})\right|\right] + \mathbb{H}(Q(\bm{\zeta} \mid \bm{\kappa})),
\end{align}
where $\bm{\zeta} = T(\cdot)$ is the one-to-one differentiable transformation, $Q(\bm{\eta} \mid \bm{\kappa})\left|\det J_{T}(\bm{\eta})\right|$ is the variational approximation in the original latent variable space, and $\mathbb{H}(\cdot)$ is the entropy term.

\begin{figure}
    \centering
    \includegraphics[width=0.99\linewidth]{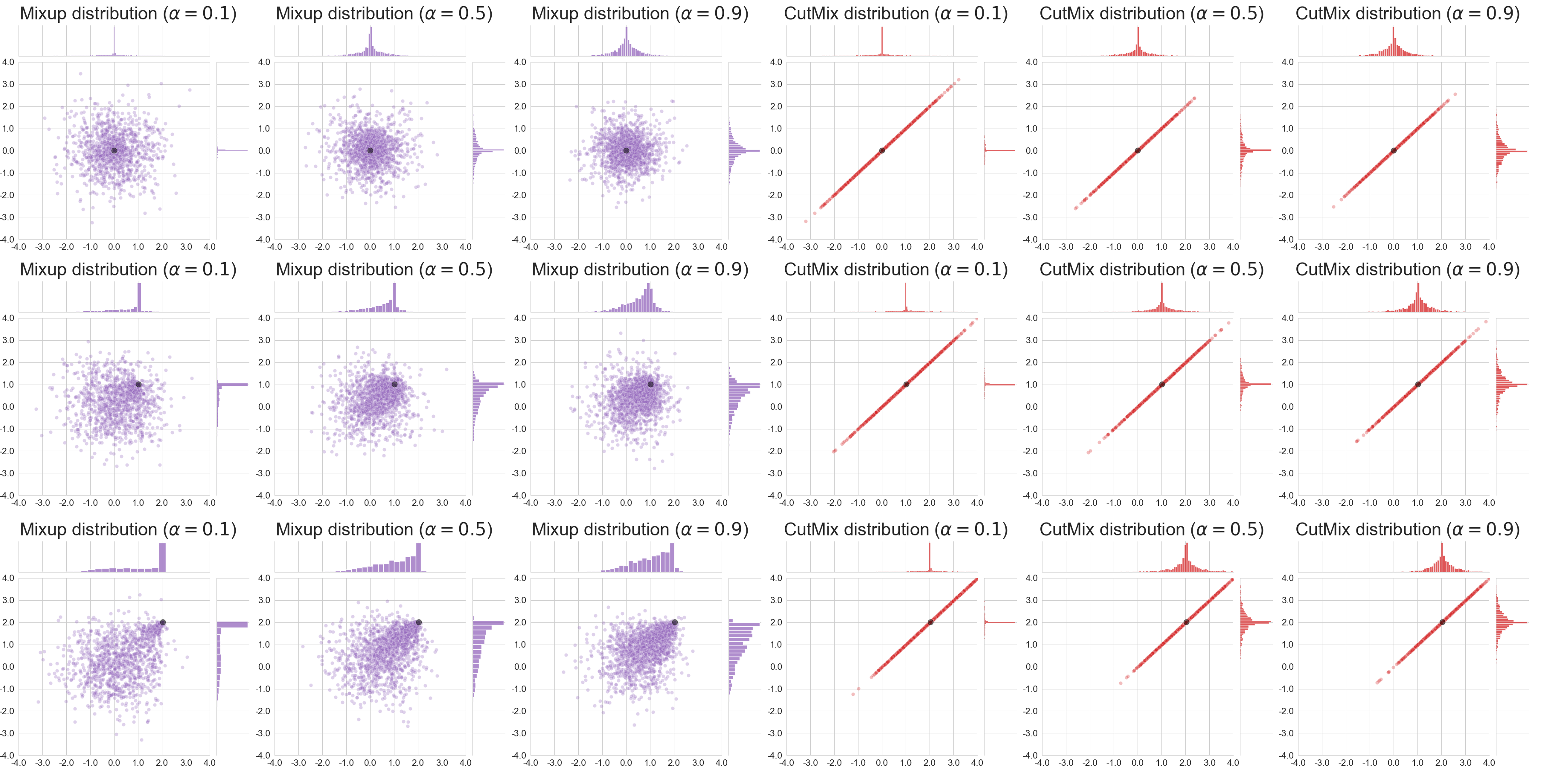}
    \caption{Plots of the distributions of points induced by mixup and cutmix (Gaussian distribution case). The black dots represent the input $\bm{x}$, and the figure shows the distributions induced by $\psi_M(\bm{x})$ and $\psi_C(\bm{x})$ when those $\bm{x}$ are fixed.}
    \label{fig:mixup_distributions}
\end{figure}

\begin{figure}
    \centering
    \includegraphics[width=0.99\linewidth]{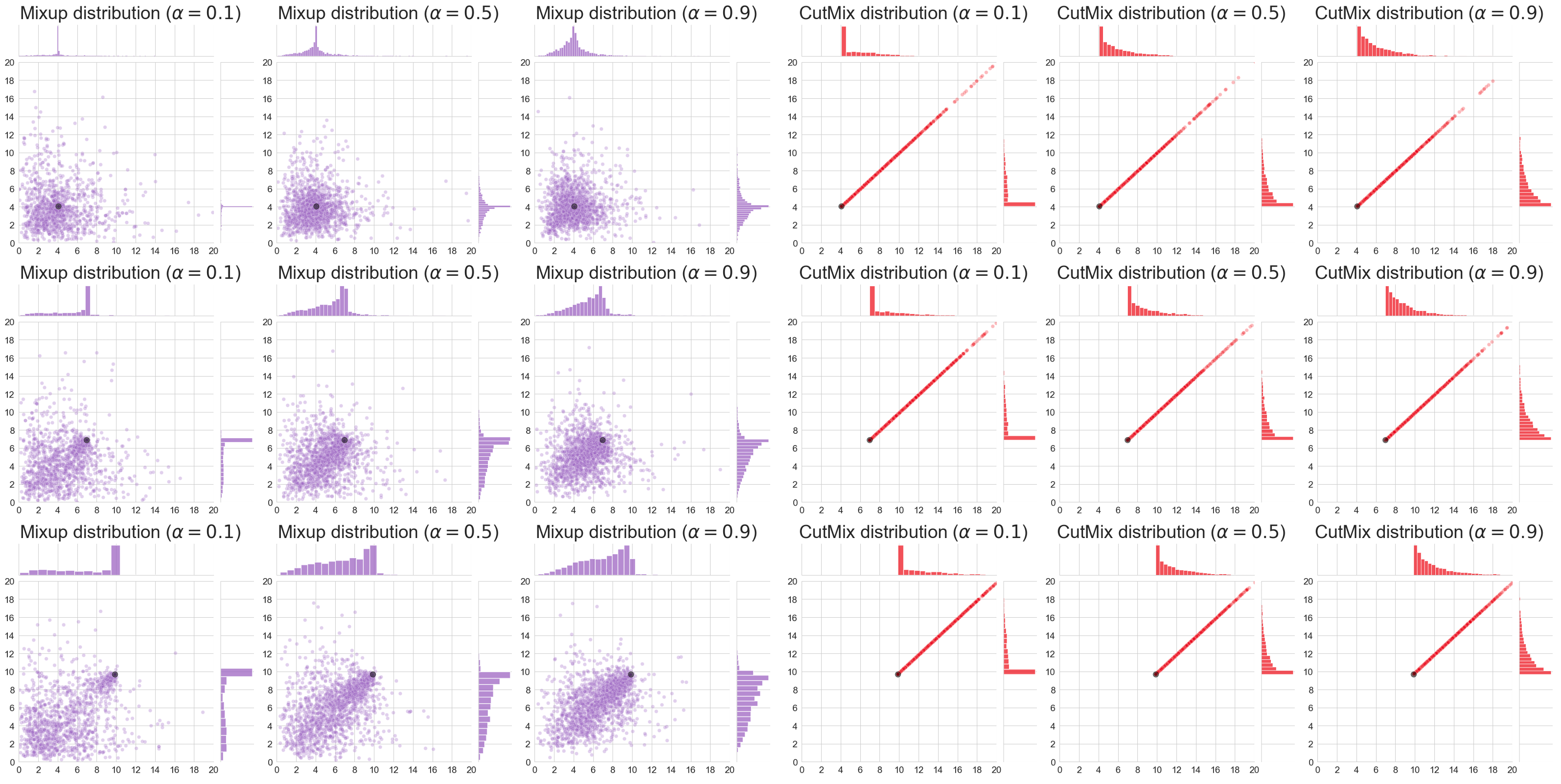}
    \caption{Plots of the distributions of points induced by mixup and cutmix (Gamma distribution case). The black dots represent the input $\bm{x}$, and the figure shows the distributions induced by $\psi_M(\bm{x})$ and $\psi_C(\bm{x})$ when those $\bm{x}$ are fixed.}
    \label{fig:mixup_distributions_gamma}
\end{figure}

\section{Numerical Experiments}
\label{sec:numerical_experiments}
In this section, we conduct several numerical experiments for our VB-TTA framework.
In our implementation, we use the PyMC 5.16.1~\citep{pymc2023} for the variational inference, PyTorch 2.3.1~\citep{NEURIPS2019_9015} for the neural networks, Numpy 2.0.0~\citep{harris2020array} for the linear algebra and matrix operations, and matplotlib 3.9.1~\citep{Hunter:2007} for plots and figures.

\subsection{Illustrative Examples}
\label{sec:illustrative_examples}
Our VB-TTA is formalized as a mixture model of the Gaussian distributions generated by the data augmentation methods and thus assumes that the distribution induced by each data augmentation is a Gaussian distribution.
Injecting Gaussian noise, one of the most commonly used data augmentations, clearly satisfies this assumption.
To expand the possible options, we further investigate how this assumption works using mixup~\citep{zhang2017mixup} and cutmix~\citep{yun2019cutmix}, two more complex data augmentation methods that have been popular in recent years.
The mixup is a data augmentation method that generates a weighted average $\tilde{\bm{x}} = (1-\lambda)\bm{x}_i + \lambda \bm{x}_j$ of two instances $\bm{x}_i, \bm{x}_j$ using $\lambda \in [0, 1]$ obtained from a Beta distribution $\mathcal{B}(\alpha, \alpha)$ for some parameter $\alpha\in (0, 1)$. 
Also, cutmix is a technique that replaces some dimensions of one instance $\bm{x}_i$ with the corresponding dimensions of another instance $\bm{x}_j$.
Consider the two-dimensional case $\bm{x}_i=(x_{i,1}, x_{i,2}), \bm{x}_j=(x_{j,1}, x_{j,2})$ and let the dimension replaced by cutmix be the second dimension, the generated instance is $\tilde{x} = (x_{i,1}, x_{j,2})$.
When mixup and cutmix are used in the TTA framework, new input is generated using input instance $\bm{x}$ at test time and instance $\bm{x}^* \in \mathcal{D}$ randomly sampled from the training data.
Let $\psi_M(\bm{x})\colon\mathcal{X}\to\mathcal{X}$ and $\psi_C(\bm{x})\colon\mathcal{X}\to\mathcal{X}$ be functions corresponding to mixup and cutmix, respectively.
\begin{align*}
    \psi_M(\bm{x}) &\coloneqq (1-\lambda)\bm{x} + \lambda\bm{x}^*, \quad\quad\quad\ \bm{x}^*\sim\mathcal{D},\quad \lambda\sim\mathcal{B}(\alpha, \alpha), \\
    \psi_C(\bm{x}) &\coloneqq \bm{M}\odot\bm{x} + \bm{M}\odot\bm{x}^*,\quad\quad \bm{x}^*\sim\mathcal{D},\quad \bm{M}\in\{\lambda_1,\dots,\lambda_d\}, \quad \lambda_i \sim\mathcal{B}(\alpha, \alpha),
\end{align*}
where $\alpha\in(0, 1)$.
Here, the original cutmix corresponds to the $\lambda = 1$ case, but in order to conduct a unified experiment with the mixup, we assume that the mask $\bm{M}$ depends on the $\lambda$ following a Beta distribution.

First, we generate artificial data with a sample size of 10,000 following $\mathcal{N}(0,I)$ on a two-dimensional plane, where $I$ is the two-dimensional identity matrix.
Then, we investigate the distributions that the instances generated by $\psi_M$ and $\psi_C$ follow when a certain new input $\bm{x}$ is fixed.
From Figure~\ref{fig:mixup_distributions}, we can see that the probability distribution that the points induced by $\psi_M$ follow is close to the Gaussian distribution when the original input $\bm{x}$ is close to the mean of the distribution of the training data, and the further $\bm{x}$ moves to the tail of the data distribution, the further away from the Gaussian distribution it is.
In contrast, the distribution that the points induced by $\psi_C$ follow is close to the Gaussian distribution, regardless of the original input points.
In addition, Figure~\ref{fig:mixup_distributions_gamma} shows a plot of the same settings when the instances are generated from a gamma distribution as $\bm{x} \sim \mathrm{Gamma}(\alpha_\Gamma, \beta_\Gamma)$ with $\alpha_\Gamma = \beta_\Gamma = 2$.
From the above results, we can conjecture that If the inputs are close to the mean of the training data distribution, VB-TTA behaves well.

\begin{figure}
    \centering
    \includegraphics[width=0.99\linewidth]{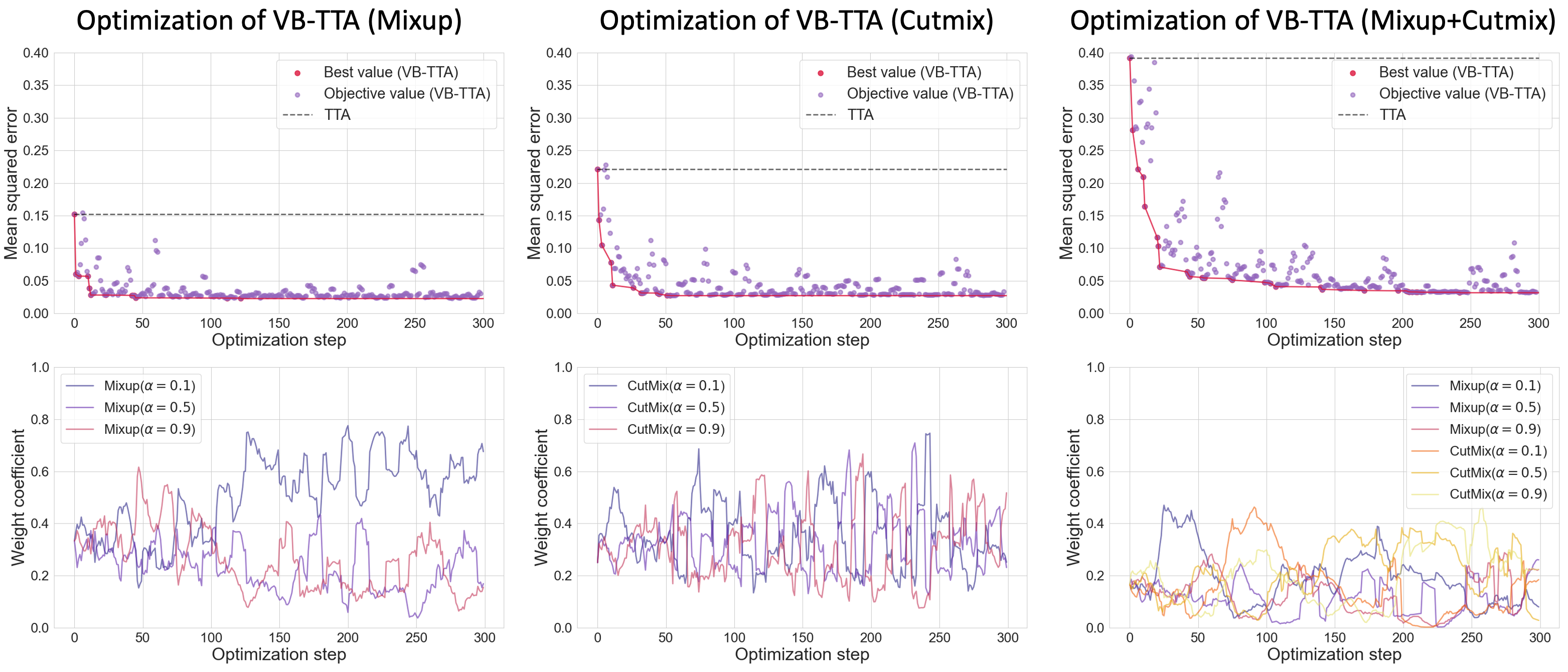}
    \caption{Optimization of VB-TTA. The first row shows the history of the optimization of the weight coefficients. The second row shows the evolution of the weight coefficients assigned to each data augmentation during the optimization process.}
    \label{fig:vb_tta_optimization}
\end{figure}

Next, we consider the numerical experiments on the artificial data.
For the supervised learning task, we generate training data with a sample size of $1,000$ following $\mathcal{N}(0, I_d)$, where $I_d$ is the $d$-dimensional identity matrix.
In this experiment, we set $d=40$.
The output labels corresponding to the input instances are generated according to a random polynomial with noise for each trial.
To induce a noisy training environment, we added different labels with noise to 30\% of all instances.
As a base model, we use a three-layer multilayer perceptron with relu as the activation function, and the mean square error as the evaluation metric.
We consider mixup and cutmix with $\alpha \in \{0.1, 0.5, 0.9\}$ as data augmentation methods used in the TTA framework.
We optimize the weight coefficients in $300$ steps by VB-TTA.
Table~\ref{tab:toy_data_experiments} shows the experimental results on the artificial dataset.
In this table, $K$-TTA stands for the $K$ data augmentations, each of which has a weight factor of $1/K$.
In addition, $K$-VB-TTA is our framework for optimizing the weight coefficients for each of the $K$ data augmentations.
Table~\ref{tab:toy_data_experiments_gamma} also shows the experimental results of the Gamma distribution case.
From these tables, we can see that the prediction performance improves with incremental optimization of the VB-TTA.
It also shows that sufficiently good weighting coefficients are obtained in the early stages of optimization.
Furthermore, the larger the number of candidate data augmentations $K$, the worse the prediction performance when the weights are not optimized, while the better the final performance achieved by optimizing the weight coefficients.
The same trend can be seen in the first row of Figure~\ref{fig:vb_tta_optimization}, where we see a trade-off between the cost of optimizing the weight coefficients and the predictive performance that can be achieved.
The second row of Figure~\ref{fig:vb_tta_optimization} also shows the evolution of the weight coefficients assigned to each data augmentation during the optimization process.
This figure indicates that for the set of mixup augmentations, a large weight factor tends to be assigned to one of them, while for the set of cutmix augmentations, all candidates tend to be given close weights.

\begin{table}[t]
    \centering
    \caption{Experimental results on the artificial data (Gaussian case). The evaluation metric is the mean squared error, and the means and standard deviations of ten trials with different random seeds are reported.}
    \label{tab:toy_data_experiments}
    \resizebox{\columnwidth}{!}{
    \begin{tabular}{c|lllll}
        \toprule
        Strategy & Step 1 & Step 50 & Step 100 & Step 200 & Step 300 \\
        \hline
        $3$-TTA (Mixup) & $0.147 (\pm 0.044)$ & - & - & - & - \\
        $3$-TTA (Cutmix) & $0.201 (\pm 0.060)$ & - & - & - & - \\
        $6$-TTA (Mixup + Cutmix) & $0.381 (\pm 0.039)$ & - & - & - & - \\ \midrule
        $3$-VB-TTA (Mixup) & $0.147 (\pm 0.044)$ & $0.040 (\pm 0.019)$ & $0.039 (\pm 0.013)$ & $0.038 (\pm 0.014)$ & $0.038 (\pm 0.012)$ \\
        $3$-VB-TTA (Cutmix) & $0.201 (\pm 0.060)$ & $0.062 (\pm 0.028)$ & $0.039 (\pm 0.024)$ & $0.037 (\pm 0.013)$ & $0.033 (\pm 0.012)$ \\
        $6$-VB-TTA (Mixup + Cutmix) & $0.381 (\pm 0.039)$ & $0.087 (\pm 0.020)$ & $0.048 (\pm 0.016)$ & $0.035 (\pm 0.011)$ & $0.026 (\pm 0.011)$ \\
        \bottomrule
    \end{tabular}
    }
\end{table}
\begin{table}[t]
    \centering
    \caption{Experimental results on the artificial data (Gamma case). The evaluation metric is the mean squared error, and the means and standard deviations of ten trials with different random seeds are reported.}
    \label{tab:toy_data_experiments_gamma}
    \resizebox{\columnwidth}{!}{
    \begin{tabular}{c|lllll}
    \toprule
    Strategy & Step 1 & Step 50 & Step 100 & Step 200 & Step 300 \\
    \hline
    $3$-TTA (Mixup) & $0.176 (\pm 0.075)$ & - & - & - & -\\
    $3$-TTA (Cutmix) & $0.230 (\pm 0.103)$ & - & - & - & -\\
    $6$-TTA (Mixup + Cutmix) & $0.385 (\pm 0.066)$ & - & - & - & -\\
    \midrule
    $3$-VB-TTA (Mixup) & $0.176 (\pm 0.075)$ & $0.137 (\pm 0.052)$ & $0.088 (\pm 0.025)$ & $0.052 (\pm 0.019)$ & $0.043 (\pm 0.017)$ \\
    $3$-VB-TTA (Cutmix) & $0.230 (\pm 0.103)$ & $0.081 (\pm 0.079)$ & $0.040 (\pm 0.024)$ & $0.038 (\pm 0.015)$ & $0.033 (\pm 0.013)$\\
    $6$-VB-TTA (Mixup + Cutmix) & $0.385 (\pm 0.066)$ & $0.099 (\pm 0.024)$ & $0.053 (\pm 0.020)$ & $0.043 (\pm 0.014)$ & $0.029 (\pm 0.013)$ \\
    \bottomrule
    \end{tabular}
    }
\end{table}

\subsection{Experimental Results on Real Datasets}
\label{sec:real_datasets}
Finally, we consider investigating the behavior of our framework on real datasets.
We conduct our experiments on the following datasets.
\begin{itemize}
    \item {\bf CIFAR-10 N}~\citep{wei2021learning}: CIFAR-10 N dataset includes the training dataset of CIFAR-10 with human-annotated real-world noisy labels collected from Amazon Mechanical Turk. For these datasets, each training image contains one clean label and three human-annotated labels.
human-annotated labels.
    \item {\bf Food-101}~\citep{bossard2014food}: The Food-101 dataset consists of 101 food categories with 750 training and 250 test images per category, making a total of 101k images. The labels for the test images have been manually cleaned, while the training set contains some noise.
    \item {\bf UTKFace}~\citep{zhang2017age}: The dataset consists of 24,108 face images with annotations of age (ranging from 0 to 116 years old). For noisy environments, all instances in the training data are given an additional label according to $\mathcal{N}(y, 5)$.
\end{itemize}

\begin{table}[t]
    \centering
    \caption{Experimental results on the real datasets. The evaluation metric is the accuracy for CIFAR-10 N and Food-101, and the mean absolute error for UTKFace.}
    \label{tab:real_data_experiments}
    \begin{tabular}{c|lll}
        \toprule
        Strategy & CIFAR-10 N ($\uparrow$) & Food-101 ($\uparrow$) & UTKFace ($\downarrow$) \\
        \hline
        $3$-TTA (Mixup) &  $0.887 (\pm 0.013)$ & $0.761 (\pm 0.025)$ & $7.820 (\pm 2.103)$ \\
        $3$-TTA (CutMix) &  $0.895 (\pm 0.014)$ & $0.755 (\pm 0.016)$ & $7.451 (\pm 1.640)$ \\
        $6$-TTA (Mixup + CutMix) &  $0.909 (\pm 0.010)$ & $0.783 (\pm 0.022)$ & $7.409 (\pm 1.812)$ \\ \midrule
        $3$-VB-TTA (Mixup) & $0.916 (\pm 0.011)$ & $0.779 (\pm 0.024)$ & $7.001 (\pm 0.625)$ \\
        $3$-VB-TTA (CutMix) &  $0.921 (\pm 0.009)$ & $0.773 (\pm 0.015)$ & $6.688 (\pm 0.757)$ \\
        $6$-VB-TTA (Mixup + CutMix) &  $0.929 (\pm 0.006)$ & $0.798 (\pm 0.010)$ & $6.405 (\pm 0.491)$ \\
        \bottomrule
    \end{tabular}
\end{table}

\begin{figure}
    \centering
    \includegraphics[width=\linewidth]{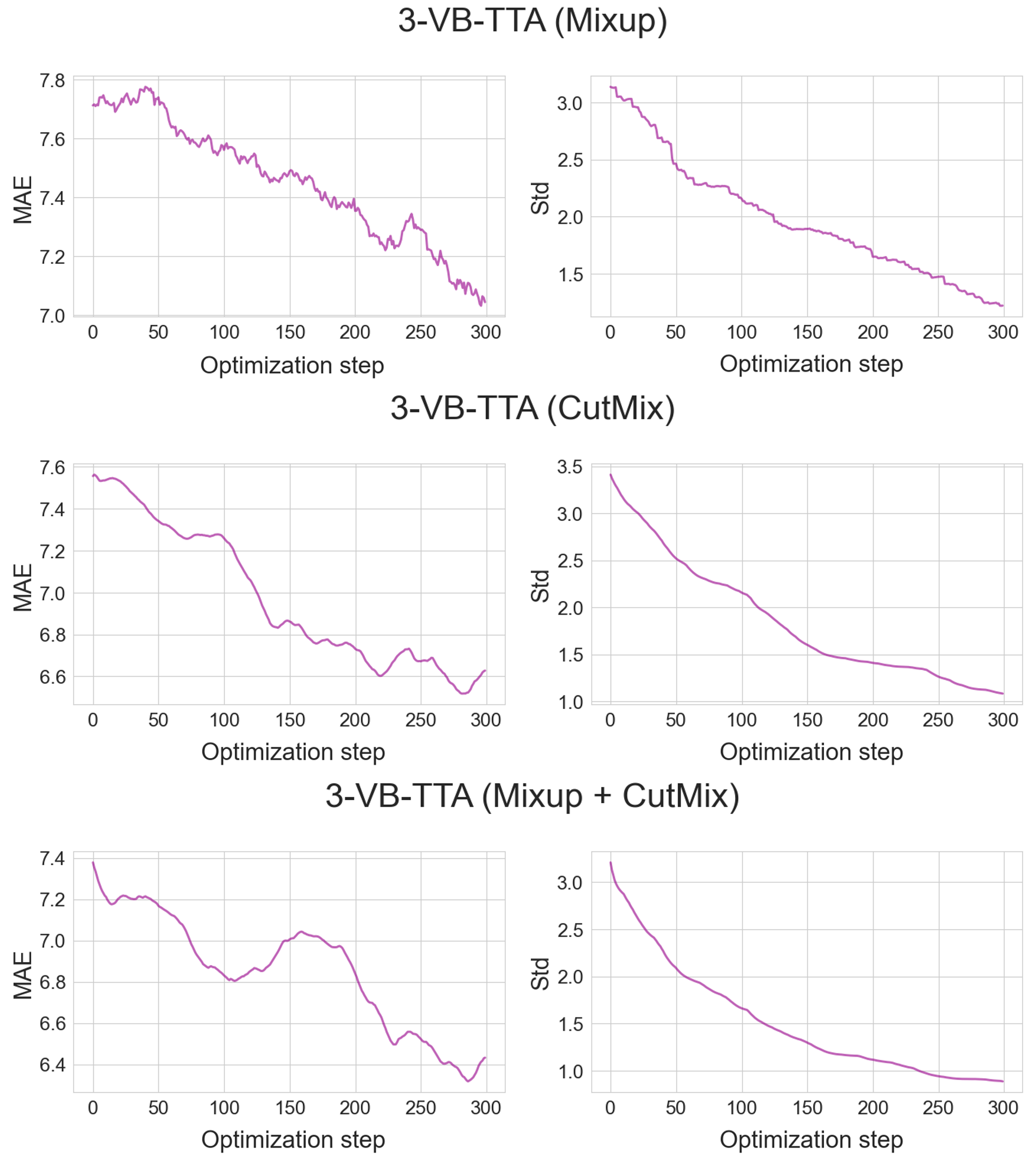}
    \caption{Tracking of the mean absolute error and the standard deviation throughout the optimization in VB-TTA.}
    \label{fig:mae_std_tracking}
\end{figure}

\begin{figure}[]
    \centering
    \includegraphics[width=0.85\linewidth]{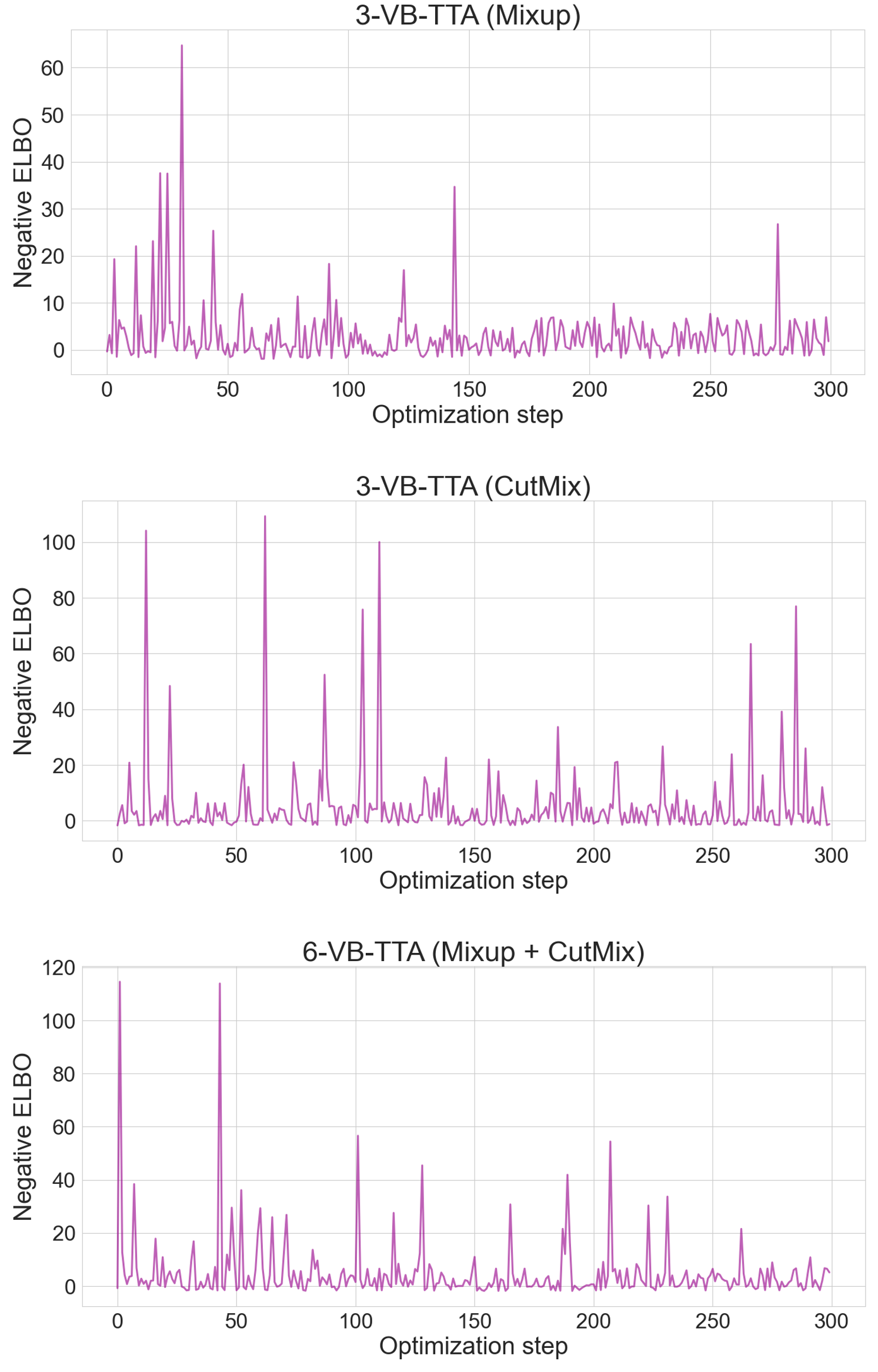}
    \caption{Tracking of the negative ELBO throughout the optimization in VB-TTA.}
    \label{fig:elbo_tracking}
\end{figure}

As a base model, we use the ResNet-18~\citep{he2016deep}.
We use accuracy as the evaluation metric for classification problems (CIFAR-10 N and Food-101) and MAE as the evaluation metric for regression problems (UTKFace).
\begin{align*}
    \mathrm{Accuracy} &\coloneqq \frac{\mathrm{\text{\# Correct Predictions}}}{\text{\# All Predictions}} = \frac{1}{N}\sum^N_{i=1}\mathbbm{1}_{\{\hat{y}_i = y_i\}}, \\
    \mathrm{MAE} &\coloneqq \frac{1}{N}\sum^N_{i=1} \left| \hat{y}_i - y_i\right|.
\end{align*}
Also, we use Adam~\citep{kingma2014adam} with $\mathrm{lr}=0.001$ as the optimizer for updating the parameters.
\begin{align*}
    \bm{\theta}_{t+1} &= \bm{\theta}_t - \mathrm{lr}\cdot\frac{\hat{\bm{m}}_t}{\sqrt{\hat{\bm{v}}} + \varepsilon}, \\
    \hat{\bm{m}}_t &\coloneqq \text{Aggregate of gradients at time $t$}, \\
    \hat{\bm{v}} &\coloneqq \text{Sum of square of past gradients}, \\
    \varepsilon &\coloneqq \text{A small positive constant}.
\end{align*}
We set the number of epochs to $200$, and the number of optimization steps for the weight coefficients to $300$ for all tasks.

Table~\ref{tab:real_data_experiments} shows the results of numerical experiments on real datasets.
We can see that optimization of the weight coefficients improves performance compared to uniformly weighted TTA.
Also, when increasing the number $K$ of data augmentations from 3-TTA to 6-TTA, the performance improvement is greater when $K$ is increased under carefully chosen weights than when $K$ is increased with uniform weights.
For example, on the UTKFace dataset, the performance improvement for 3-TTA (CutMix) $\to$ 6-TTA (Mixup + CutMix) is $7.501 - 7.488 = 0.013$, while the performance improvement for 3-VB-TTA (CutMix) $\to$ 6-VB-TTA (Mixup + CutMix) is $6.710 - 6.329 = 0.381$.
We also see that optimizing the weight coefficients reduces the variance of estimations.
This can be expected because the weight coefficients of candidates that are unnecessarily noisy in the prediction are reduced.
Finally, Figures~\ref{fig:mae_std_tracking} and~\ref{fig:elbo_tracking} show the tracking of the MAE, the standard deviation, and the negative ELBO throughout the optimization for UTKFace dataset, and one can see that the convergence of the ELBO.

\section{Conclusion and Discussion}
\label{sec:conclusion}
In this study, we considered a framework of the weighted Test-Time Augmentation (TTA) in noisy training environments.
One of the key challenges of this framework is, it is not easy to determine the weighting coefficients as seen in Section~\ref{sec:difficulty_determination_tta_weight}.
This suggests that the policy of simply excluding inaccurate data augmentation methods is not optimal.
To tackle this problem, we demonstrated that the TTA procedure can be formalized as a Bayesian mixture model by assuming that the transformed instances with each data augmentation used in TTA follow some probability distribution.
Analytically, since the formalization needs to be adjusted depending on whether the output of the predictor is continuous or discrete, these formalizations are presented in separate sections.
The use of AVDI was also considered for a uniform and reasonable implementation.
We indicated that optimization of the mixing coefficients through this formalization can suppress unnecessary candidates for data augmentation.
Numerical experiments revealed that our formalization behaves well, obtaining good weighting coefficients with a reasonable number of optimization steps.
In the illustrative example (in Section~\ref{sec:illustrative_examples}), we demonstrated how the weight coefficients evolved in our VB-TTA and showed that the optimization worked well.
Finally, numerical experiments on real-world datasets (in Section~\ref{sec:real_datasets}) show that the proposed method behaves well in the problem setting we have in mind.

To the best of our knowledge, this is the first study to combine TTA and variational Bayes, and it is hoped that this study will serve as a bridge between these two frameworks.
In particular, providing a variational Bayesian formalization of TTA suggests that the knowledge for optimization and the assets of theoretical analysis developed in the latter could be used to improve the algorithm.
Examples could include findings on prior distributions, which are frequently discussed in the Bayesian inference literature, and mathematical techniques for improving variational lower bounds.
That is, possible future works are as follows.
\begin{itemize}
    \item {\bf Non-Gaussian prior}. In this study, we assumed that the instances obtained by each data augmentation follow Gaussian distribution. However, the data augmentations used in practice are expected to induce a wide variety of distributions. Therefore, the introduction of a better prior distribution is worth considering.
    \item {\bf Improved variational lower bounds}. Our objective function is based on the maximization of the variational lower bound, which is equivalent to the minimization of the KL-divergence. In the field of variational inference, better variational lower bounds have been investigated~\citep{bamler2019tightening,yang2020alpha,zalman2022variational}. Our optimization could be improved by leveraging these known assets in variational inference.
    \item {\bf Other training assumptions}. For example, it has been reported that TTA is effective under the distribution shift assumption that training and test data follow different distributions~\citep{zhang2022memo}. It would be beneficial to have a discussion in such cases.
\end{itemize}

\vskip 0.2in
\bibliography{main}

\end{document}